\documentclass[letterpaper]{article} 
\usepackage{main}  
\usepackage{times}  
\usepackage{helvet}  
\usepackage{courier}  
\usepackage[hyphens]{url}  
\usepackage{graphicx} 
\usepackage{natbib}  
\usepackage{caption} 
\usepackage{float}
\usepackage{amsmath}
\usepackage{amsfonts}
\usepackage{amssymb}
\usepackage{booktabs}

\usepackage{algorithm}
\usepackage{algorithmic}

\usepackage{newfloat}
\usepackage{listings}
\DeclareCaptionStyle{ruled}{labelfont=normalfont,labelsep=colon,strut=off} 
\floatstyle{ruled}
\newfloat{listing}{tb}{lst}{}
\floatname{listing}{Listing}
\title{Hand1000: Generating Realistic Hands from Text with Only 1,000 Images}
\author{
    Haozhuo Zhang\textsuperscript{\rm 1},
    Bin Zhu\textsuperscript{\rm 2}\thanks{Corresponding author.},
    Yu Cao\textsuperscript{\rm 2},
    Yanbin Hao\textsuperscript{\rm 3}
}
\affiliations{
    \textsuperscript{\rm 1}Peking University\\
    \textsuperscript{\rm 2}Singapore Management University\\
    \textsuperscript{\rm 3}University of Science and Technology of China\\
    2100013132@stu.pku.edu.cn, binzhu@smu.edu.sg, yu.cao.2022@msc.smu.edu.sg, haoyanbin@hotmail.com
}

\usepackage{bibentry}

\begin{document}

\maketitle

\begin{abstract}
Text-to-image generation models have achieved remarkable advancements in recent years, aiming to produce realistic images from textual descriptions. However, these models often struggle with generating anatomically accurate representations of human hands. The resulting images frequently exhibit issues such as incorrect numbers of fingers, unnatural twisting or interlacing of fingers, or blurred and indistinct hands. 
These issues stem from the inherent complexity of hand structures and the difficulty in aligning textual descriptions with precise visual depictions of hands.
To address these challenges,  we propose a novel approach named \textbf{Hand1000} that enables the generation of realistic hand images with target gesture using only 1,000 training samples. The training of Hand1000 is divided into three stages with the first stage aiming to enhance the model’s understanding of hand anatomy by using a pre-trained hand gesture recognition model to extract gesture representation. The second stage further optimizes text embedding by incorporating the extracted hand gesture representation, to improve alignment between the textual descriptions and the generated hand images. The third stage utilizes the optimized embedding to fine-tune the Stable Diffusion model to generate realistic hand images.
In addition, we construct the first publicly available dataset specifically designed for text-to-hand image generation. Based on the existing hand gesture recognition dataset, we adopt advanced image captioning models and LLaMA3 to generate high-quality textual descriptions enriched with detailed gesture information. Extensive experiments demonstrate that Hand1000 significantly outperforms existing models in producing anatomically correct hand images while faithfully representing other details in the text, such as faces, clothing and colors.
Additional details and resources are available on our project page: \url{https://haozhuo-zhang.github.io/Hand1000-project-page/}.


\end{abstract}


\section{Introduction}
\label{sec:intro}

\begin{figure}[t]
  \centering
   \includegraphics[width=0.9\linewidth]{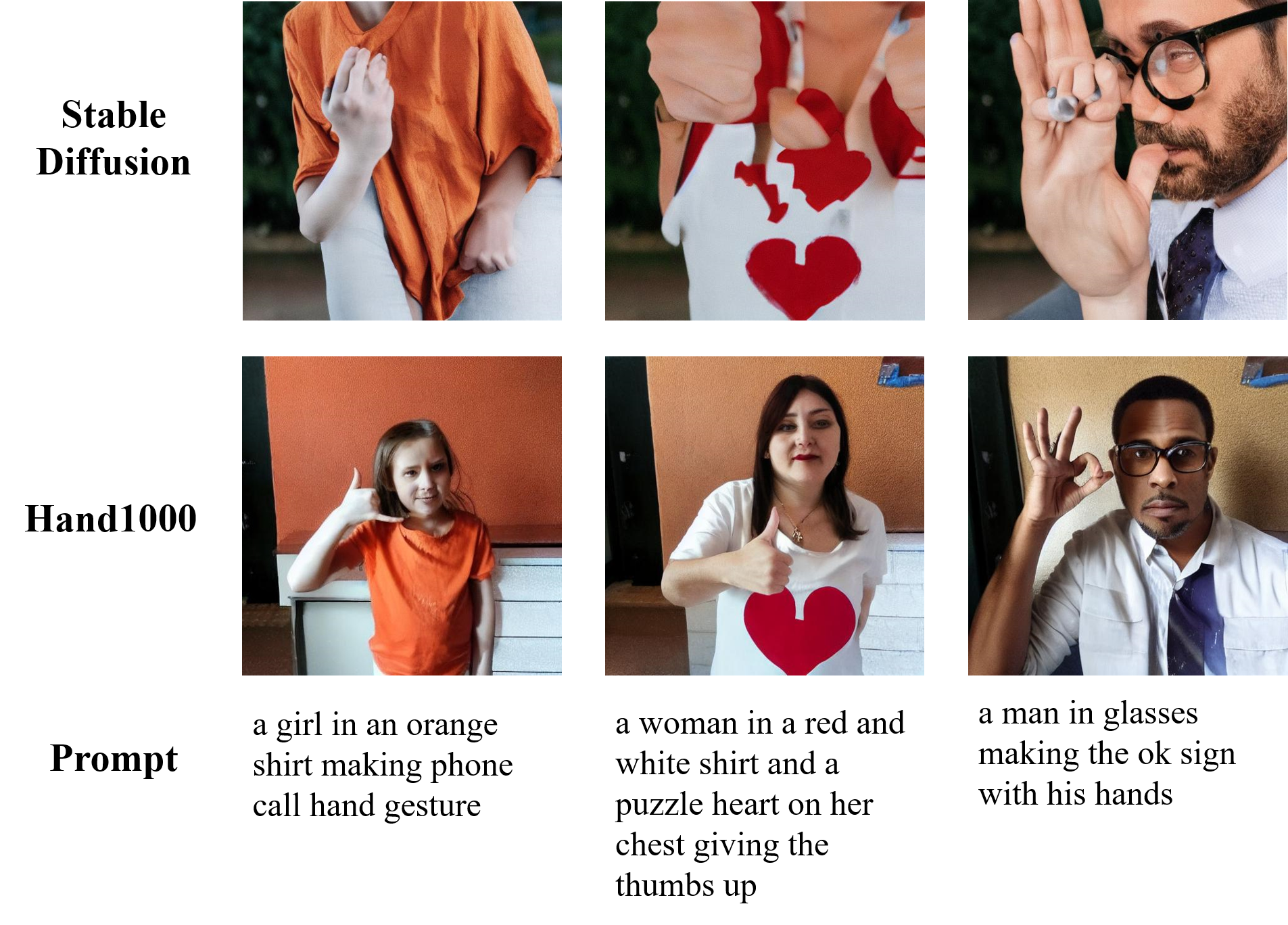}
   \caption{Comparison of hand image generation results between Stable Diffusion and our Hand1000. Given the same text prompt, Stable Diffusion produces deformed and chaotic hands. In contrast, our proposed Hand1000 manages to generate anatomically correct and realistic hands while preserving details such as character, clothing, and colors.}
   \label{fig:first_fig}
\end{figure}

In recent years, text-to-image generation models have achieved significant progress, such as Stable Diffusion~\cite{rombach2022high, esser2024scaling}, Multimodal Large Language Model (MM-LLM)~\cite{koh2024generating}, and DiT~\cite{peebles2023scalable}. However, these models often struggle when generating images involving human bodies, particularly hands. The generated images frequently exhibit distorted or misaligned body parts that do not conform to normal physical characteristics. As shown in the top of Figure \ref{fig:first_fig}, this problem is especially pronounced with hands, which often appear blurry, with fingers twisted or interlaced, and an incorrect number of fingers. Despite capturing the color and texture of human hands, these generated images clearly display noticeable inaccuracies.

The underlying causes of this phenomenon can be attributed to two primary factors: (a) The inherent complexity of the human hand's structure, along with the occlusions between fingers, means that the Stable Diffusion model lacks the necessary prior knowledge of the precise physical structure of human hands, resulting in its inability to generate anatomically correct hands. (b) When the text prompt includes information related to human hands (e.g., ``thumbs up" and ``phone call gesture"), the model struggles to associate this with corresponding hand images, indicating its inability to accurately comprehend text containing hand-related information and thus failing to produce realistic hand visual appearance.

Recent studies have explored methods to generate anatomically correct hands using diffusion models, such as ~\cite{narasimhaswamy2024handiffuser} and ~\cite{lu2023handrefiner}, yet they fail to fully address the fundamental issues outlined in the previous paragraph. Furthermore, these approaches require extensive training datasets, often comprising hundreds of thousands of images, which significantly reduces training efficiency and limits their practical applicability. In contrast, recent advancements in text-based image editing techniques~\cite{kawar2023imagic, lin2024make, bar2022text2live} have demonstrated the ability to achieve background alterations or modifications in facial expressions and actions with only a few images. 

Inspired by text-based image editing works, we propose a novel method named Hand1000 to generate anatomically correct hands using only 1,000 images for each target hand gesture in three stages.
To address the issue (a) from the previous paragraph, our proposed method enhances hand prior knowledge about correct hand anatomy by leveraging a pre-trained hand gesture recognition model (i.e., Mediapipe hands~\cite{zhang2020mediapipe}) to extract gesture representation in the first stage. To tackle issue (b), in the second stage, we optimize the text embedding for each image by integrating the hand gesture representation in the first stage to ensure alignment between text and hand image. This optimized embedding is then used to fine-tune the stable diffusion model to generate realistic hand images in the third stage. In addition, we construct a dataset specifically for text-to-hand-image generation by leveraging hand gesture recognition datasets (i.e. HaGRID dataset~\cite{kapitanov2024hagrid}). Image captioning models(e.g. BLIP2~\cite{li2023blip}, PaliGemma~\cite{beyer2024paligemma}, and VitGpt2~\cite{mishra2024image}) are first used to produce a textual description, and LLaMA3 model~\cite{touvron2023open} is employed to enrich the textual description with hand gesture information.
After training on 1,000 images in our dataset, our Hand1000 significantly outperforms the Stable Diffusion model in generating correct hands for the same textual prompts, which is shown at the bottom of Figure \ref{fig:first_fig}.
\par
In summary, the contributions of this paper are as follows: 
\begin{itemize}
    \item We propose a new method, Hand1000, that empowers the Stable Diffusion model with the ability to generate realistic hands using only 1,000 images. To the best of our knowledge, Hand1000 is the first text-to-image model capable of achieving accurate hand generation with such a small-scale dataset.

    \item By leveraging image captioning models and the LLaMA3, we construct the first publicly available dataset specifically designed for generating hand images from textual descriptions. This dataset aims to facilitate and advance research within the community on this task.

    \item We showcase the power of enhancing hand features and text optimization to generate anatomically correct and realistic hand images from text. Hand1000 achieves significant improvements compared with existing methods.
\end{itemize}
\section{Related Work}
\label{sec:Related Work}

\subsection{Text-to-Image Generation}

Text-based image generation has long been a prominent research area, with classical methods including Generative Adversarial Networks (GANs) and their variants~\cite{kang2023scaling, tao2022df, xu2018attngan, zhu2020cookgan, ye2021improving, zhu2019dm, zhu2019r2gan, wang2024consistency, li2023reganie, ma2023cfft}, as well as stable diffusion models and their derivatives~\cite{sauer2023adversarial, saharia2022photorealistic, dhariwal2021diffusion, ho2020denoising, ho2022cascaded, sohl2015deep, yang2024fontdiffuser, ye2024diffusionedge, yu2024accelerating}. These methods leverage the capabilities of CLIP~\cite{radford2021learning} and Transformer~\cite{vaswani2017attention}, empowering them to generate images from textual descriptions. In recent years, numerous related studies have emerged, achieving notable success in the field such as GigaGAN~\cite{kang2023scaling}, Imagen~\cite{saharia2022photorealistic}, and SDXL Turbo~\cite{sauer2023adversarial}. With the advent of ChatGPT~\cite{brown2020language, achiam2023gpt} and the development of large language models(LLM)~\cite{chowdhery2023palm,touvron2023llama,team2023gemini}, several multimodal models for text-to-image generation have also emerged, such as DALL-E~\cite{ramesh2021zero}. Despite the notable advancements these methods have achieved in image generation, they frequently encounter significant issues when producing images that include human hands, often resulting in severe distortions or anomalies in hand representations.

\subsection{Realistic Hand Generation}

Text-based diffusion models are effective for generating various images but struggle with rendering human hands. Recent works address this challenge in two ways. Some studies use image inpainting, like HandRefiner~\cite{lu2023handrefiner}, which identifies and crops hand regions, then refines them using ControlNet~\cite{zhang2023adding}. However, this approach depends on another model for initial image generation and fails with severely distorted hands. Alternatively, methods like HanDiffuser~\cite{narasimhaswamy2024handiffuser} directly generate hand images by reconstructing text prompts into 3D hand parameters, though the training process is complex and expensive. Additionally, the above methods require an enormous training dataset, in the order of hundreds of thousands of samples, to achieve noticeable results. In contrast, our \textbf{Hand1000} method fine-tunes on just \textbf{1,000} images to generate realistic hands without requiring explicit hand pose modeling, offering a more efficient and flexible approach. 

\subsection{Text-based Image Editing}

Text-based image editing has seen rapid advancements in recent years, allowing for the modification of an image's background, style, subject size, and movements through direct textual input. These technologies, often derived from improvements in GANs~\cite{abdal2021styleflow, harkonen2020ganspace, lang2021explaining, patashnik2021styleclip, shen2020interpreting, shen2021closed, xia2023feditnet} or Stable Diffusion models~\cite{ruiz2023dreambooth, kawar2023imagic, nichol2021improved, rombach2022high, saharia2022palette, saharia2022image, song2020denoising, song2019generative, song2020improved}, typically require minimal input images to achieve the desired effects. For instance, Imagic~\cite{kawar2023imagic} firstly fine-tunes text embeddings, then fine-tunes the diffusion model, and finally performs interpolation. This method focuses on text embeddings, and achieves remarkable results. Considering the challenge of generating realistic hand images, it seems plausible to borrow from the aforementioned image editing concepts. However, image editing algorithms cannot be directly applied to generate realistic hand images because they fundamentally rely on the diffusion model's inherent capabilities, which lack prior knowledge of realistic hands. Besides, these methods usually take several minutes to modify a single picture, making it impossible to generate a large amount of images in a short time. To address these problems, we propose \textbf{Hand1000}, a method that fine-tunes the model using image editing principles. This approach not only imparts the model with knowledge of realistic hands but also benefits from the characteristics of image editing algorithms, requiring fewer training images and faster training speed.

\section{Method}
\label{sec:Method}

\newcounter{mycounter}

\begin{figure*}[t!]
  \centering
   \includegraphics[width=1.0\linewidth]{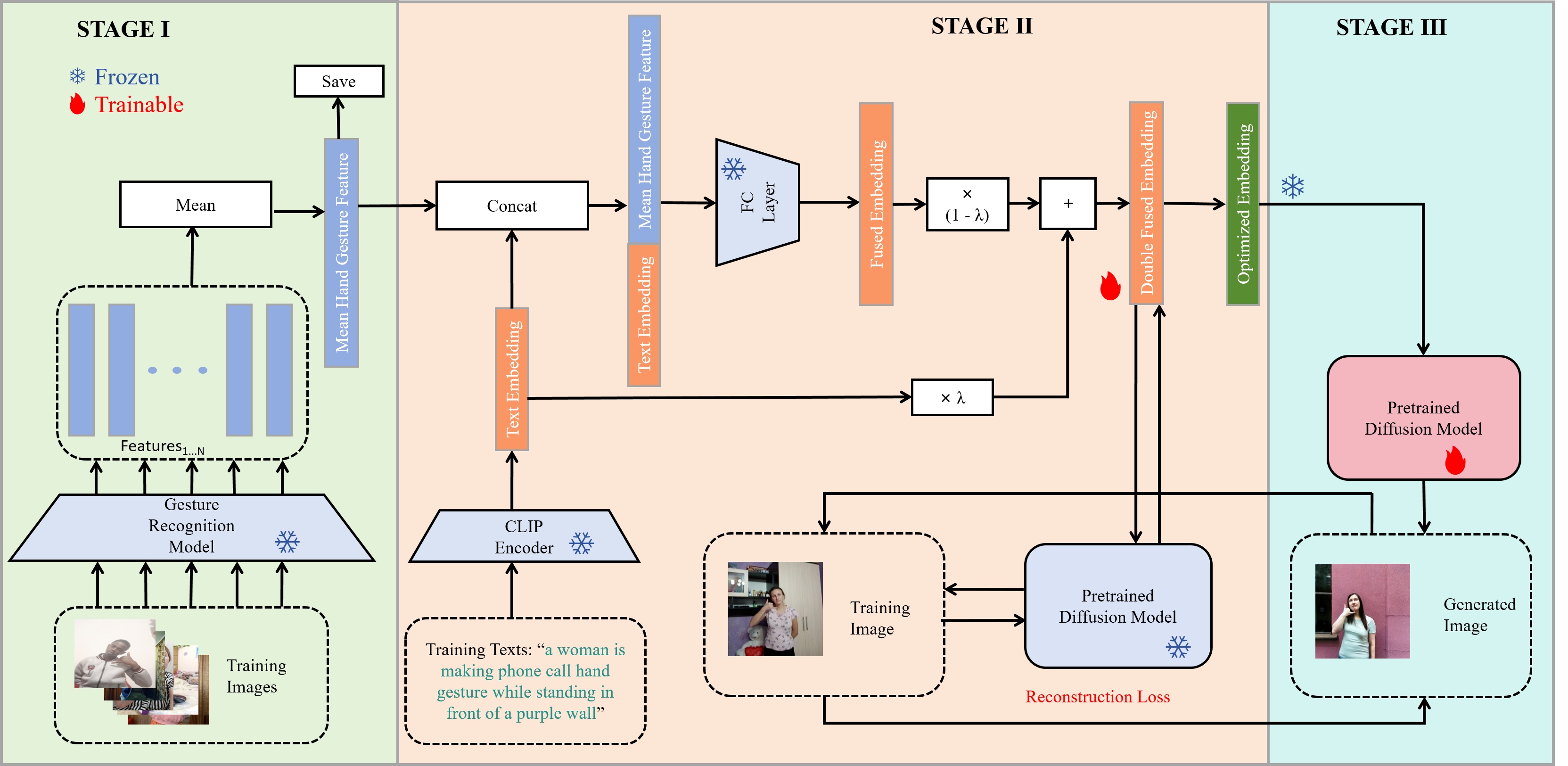}
   \caption{The proposed Hand1000 is designed with a three-stage training process. In Stage \setcounter{mycounter}{1}\Roman{mycounter}, the primary objective is to compute mean hand gesture feature from images. Stage \setcounter{mycounter}{2}\Roman{mycounter} builds on this by concatenating the mean hand gesture feature obtained in Stage \setcounter{mycounter}{1}\Roman{mycounter} with the corresponding text embeddings. These concatenated features are then mapped into a fused embedding, which is further enhanced by linearly fusing it with the original text embedding, resulting in a double-fused embedding. This embedding is optimized using a reconstruction loss through a frozen Stable Diffusion model, ensuring that the final embedding is well-optimized. Stage \setcounter{mycounter}{3}\Roman{mycounter} involves fine-tuning the Stable Diffusion model for image generation, leveraging the frozen optimized embedding obtained from Stage \setcounter{mycounter}{2}\Roman{mycounter}.}
   \label{fig:traing_fig}
\end{figure*}

This section begins by introducing the preliminaries of the text-to-image generation model, followed by a detailed explanation of the Hand1000 method. As depicted in Figure \ref{fig:traing_fig}, the training process of Hand1000 is divided into three stages. In the first stage, 1,000 images from the training set are processed by a hand gesture recognition model (i.e. Mediapipe hands~\cite{zhang2020mediapipe}) to obtain mean hand gesture feature. The second stage involves sequentially feeding text-image pairs for training, where text embedding is obtained using CLIP~\cite{radford2021learning} from text description. The text embedding is then concatenated with the mean hand gesture feature from the first stage, which is 
mapped to obtain the fused embedding. The fused embedding is linearly combined with text embedding to produce double-fused embedding, which is optimized by using the reconstruction loss with a frozen stable diffusion model to produce optimized embedding. The third stage leverages the frozen optimized embedding to further fine-tune the stable diffusion model to generate realistic hand images. 

\subsection{Preliminaries}
Text-to-image generation aims to generate an image that accurately represents a given textual description. The process of a standard text-to-image diffusion model~\cite{rombach2022high} is as follows: Given an initial noise image $\epsilon \sim \mathcal{N}(0, I)$, where the noise follows a Gaussian distribution, the CLIP text encoder~\cite{radford2021learning} $\Gamma$ encodes the \textit{text} into a text embedding $\textit{e} = \Gamma(\textit{text})$. Using \textit{e} and the time step \textit{t} as conditions, the diffusion model generates a latent representation 
$\mathbf{z}_t := \alpha_t \mathbf{x} + \sigma_t \epsilon$ based on the initial noise image $\epsilon$. Subsequently, an image is generated from this latent representation. The entire process is guided by the following mean squared error loss:
\begin{equation}
\mathbb{E}_{\mathbf{x}, \mathbf{e}, \epsilon, t} \left[ w_t \left \| \hat{\mathbf{x}}_\theta(\alpha_t \mathbf{x} + \sigma_t \epsilon, \mathbf{e}) - \mathbf{x} \right \|_2^2 \right],
\label{eq:eq1}
\end{equation}
where \( \mathbf{x} \) represents the real image, and \( \mathbf{e} \) denotes the text embedding. The terms \( \alpha_t \), \( \sigma_t \), and \( w_t \) are parameters that control the noise scheduling and sample quality, and they are functions of the diffusion process time \( t \sim U([0, 1]) \).

\subsection{Stage \setcounter{mycounter}{1}\Roman{mycounter}: Hand Gesture Feature Extraction}
Assume we have a set of images depicting a target hand gesture, such as ``phone call". To extract hand features associated with this gesture, we feed the images into a gesture recognition model(i.e. Mediapipe hands~\cite{zhang2020mediapipe}) to obtain features from the final layer of the network.
Subsequently, these features are averaged to obtain a \textbf{Mean Hand Gesture Feature} representation of the gesture, which is used for training in the following stages. Simultaneously, the Mean Hand Gesture Feature is also preserved and utilized during the inference phase.

\subsection{Stage \setcounter{mycounter}{2}\Roman{mycounter}: Text Embedding Optimization}
In the second stage of training, the main objective is to integrate text embedding with hand gesture features to facilitate the diffusion model to generate realistic hand images. Given a textual description as input, CLIP text encoder is employed to obtain the \textbf{Text Embedding}. Subsequently, the text embedding is concatenated with the \textbf{Mean Hand Gesture Feature} obtained from the first stage. 
To fuse the text and hand gesture features, a fully connected (FC) layer is employed to map the concatenated embedding to the same dimension of text embedding, resulting in the \textbf{Fused Embedding}, which smoothly incorporates the gesture's general characteristics.
Notably, the FC layer is kept frozen to reduce computational costs, as the embedding will undergo further optimization in Stage \setcounter{mycounter}{2}\Roman{mycounter}.
A linear fusion of the fused embedding with the original text embeddings is subsequently performed to obtain the \textbf{Double Fused Embedding}:
\begin{equation}
e_d = \lambda \times e_t +  \left ( 1 - \lambda  \right ) \times e_f ,
\label{eq:eq2}
\end{equation}
where \( \mathbf{\lambda} \) is the hyperparameter, \( \mathbf{e_d} \) is Double Fused Embedding, \( \mathbf{e_t} \) is Text Embedding, and \( \mathbf{e_f} \) is Fused Embedding. This linear fusion reinforces the text embedding's unique features, allowing it to leverage the gesture feature without overpowering its own nuances.

Inspired by image editing techniques~\cite{kawar2023imagic}, we proceed to optimize the Double Fused Embedding to better align with the real hand images. Specifically, 
the stable diffusion model is kept frozen while the Double Fused Embedding is fine-tuned using the loss function defined in Equation \ref{eq:eq1}, resulting in what we term \textbf{Optimized Embedding}. The Optimized Embedding, enriched and refined with hand gesture information, establishes a stronger alignment with the corresponding hand images, making it ideal for the subsequent hand image generation in Stage \setcounter{mycounter}{3}\Roman{mycounter}.

\subsection{Stage \setcounter{mycounter}{3}\Roman{mycounter}: Stable Diffusion Fine-tuning}
In the final stage of training, the Optimized Embedding obtained in Stage \setcounter{mycounter}{2}\Roman{mycounter} is used as a condition to fine-tune the Stable Diffusion model with the loss function specified in Equation \ref{eq:eq1}.
The fine-tuning procedure follows the standard text-to-image diffusion model training, with the only modification being the replacement of the usual text embedding with our Optimized Embedding from Stage II. To preserve the hand gesture information integrated in Stage II, the Optimized Embedding is kept frozen during this stage, in line with image editing works~\cite{kawar2023imagic}.

\subsection{Inference}
\begin{figure}[t!]
  \centering
   \includegraphics[width=1.0\linewidth]{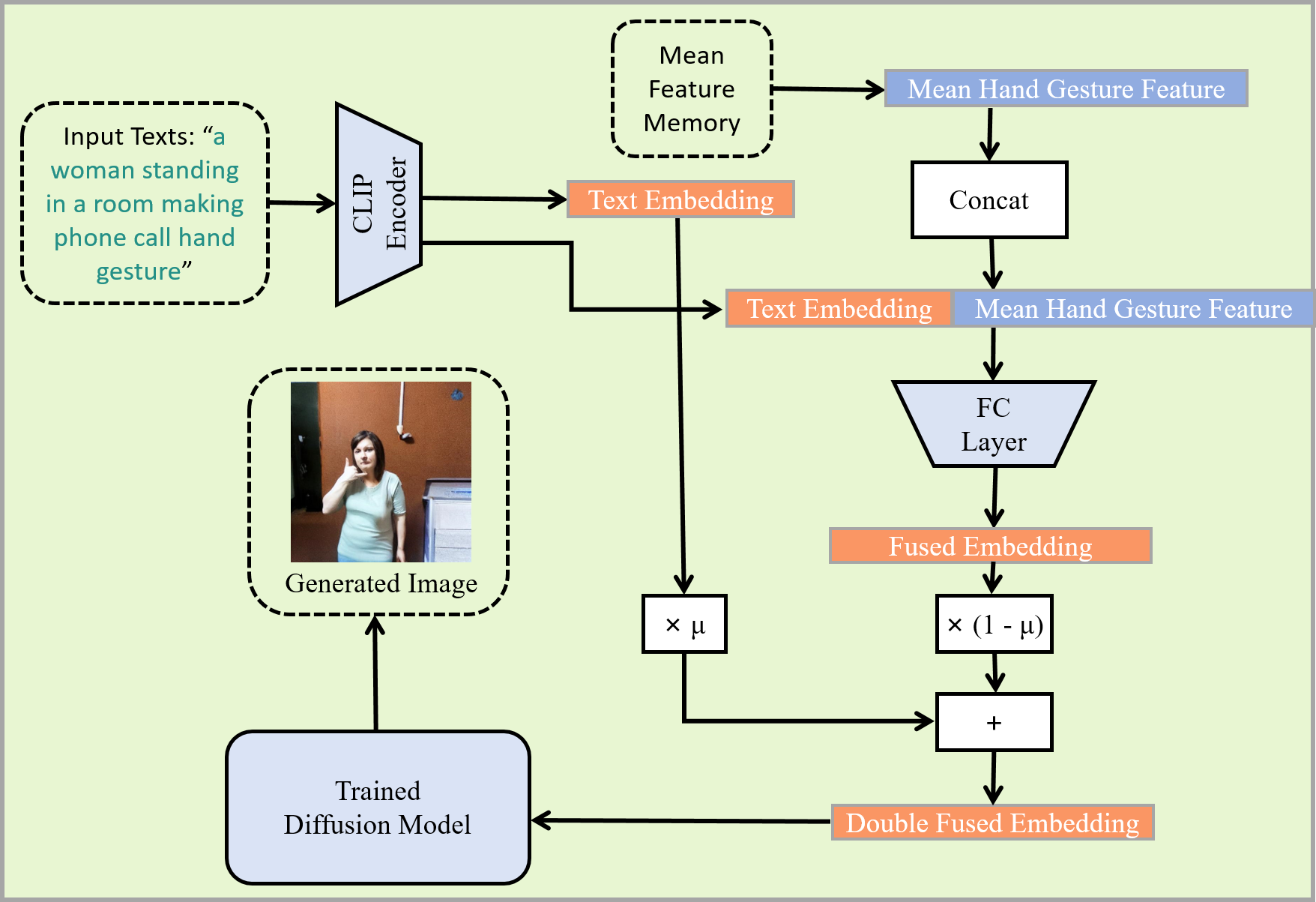}
   \caption{Overview of inference phase. First, text embedding is computed given a textual description as input. Next, the stored mean hand gesture feature is concatenated and fused with text embedding to obtain double-fused embedding. Finally, the double-fused embedding is fed into the trained diffusion model in Stage \setcounter{mycounter}{3}\Roman{mycounter} to generate hand images.}
   \label{fig:inference_fig}
\end{figure}

As illustrated in Figure \ref{fig:inference_fig}, during the inference phase, a textual description containing a hand gesture is used as input. Similar to the training stage, the CLIP text encoder is adopted to obtain the Text Embedding. The Mean Hand Gesture Feature, stored from during the Stage \setcounter{mycounter}{1}\Roman{mycounter} in the training phase, is concatenated with the Text Embedding. This concatenated feature is then reduced in dimension using a fully connected layer to obtain the Fused Embedding. 
The Text Embedding and the Fused Embedding are linearly fused to produce the Double Fused Embedding: 
\begin{equation}
e_d = \mu \times e_t +  \left ( 1 - \mu  \right ) \times e_f,
\label{eq:eq3}
\end{equation}
where \( \mathbf{\mu} \) is the hyperparameter, \( \mathbf{e_d} \) is {Double Fused Embedding}, \( \mathbf{e_t} \) is {Text Embedding}, and \( \mathbf{e_f} \) is {Fused Embedding}.The Double Fused Embedding is then used as a condition to input into the trained Stable Diffusion model in Stage \setcounter{mycounter}{3}\Roman{mycounter}, enabling the generation of images that incorporate the specified gesture and maintain a realistic hand appearance.

\section{Experiments}
\label{sec:Experiments}

\begin{figure}[t!]
  \centering
   \includegraphics[width=1.0\linewidth]{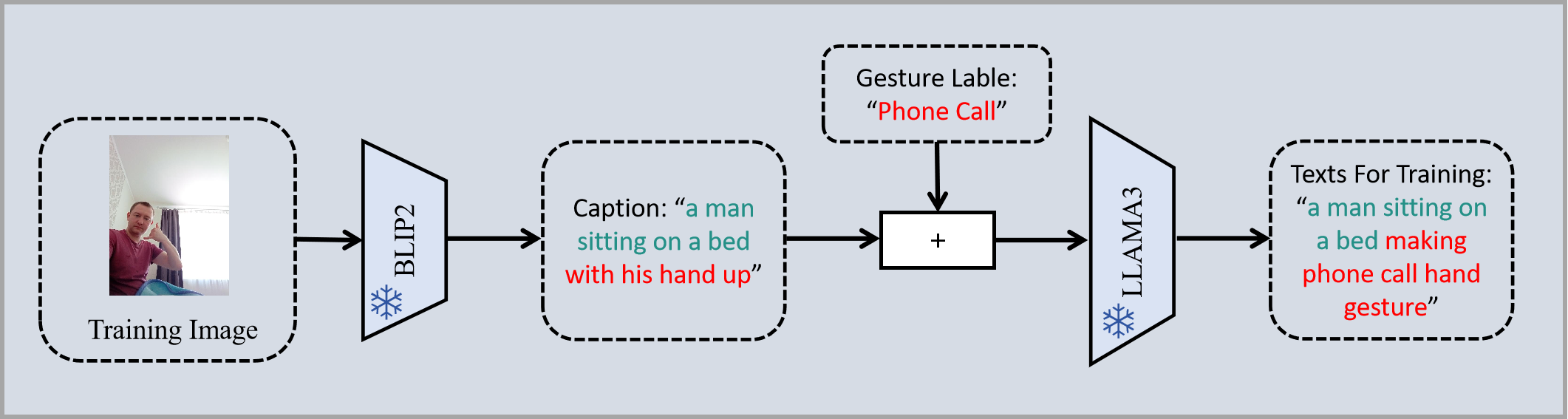}
   \caption{The dataset construction begins with generating a textual description using an image captioning model (e.g., BLIP2) from image. The textual description, along with gesture labels, is then fed into the LLaMA3 model~\cite{touvron2023open} to produce a text description enriched with gesture label information.}
   \label{fig:dataset_fig}
\end{figure}

\begin{figure}[t!]
  \centering
   \includegraphics[width=0.9\linewidth]{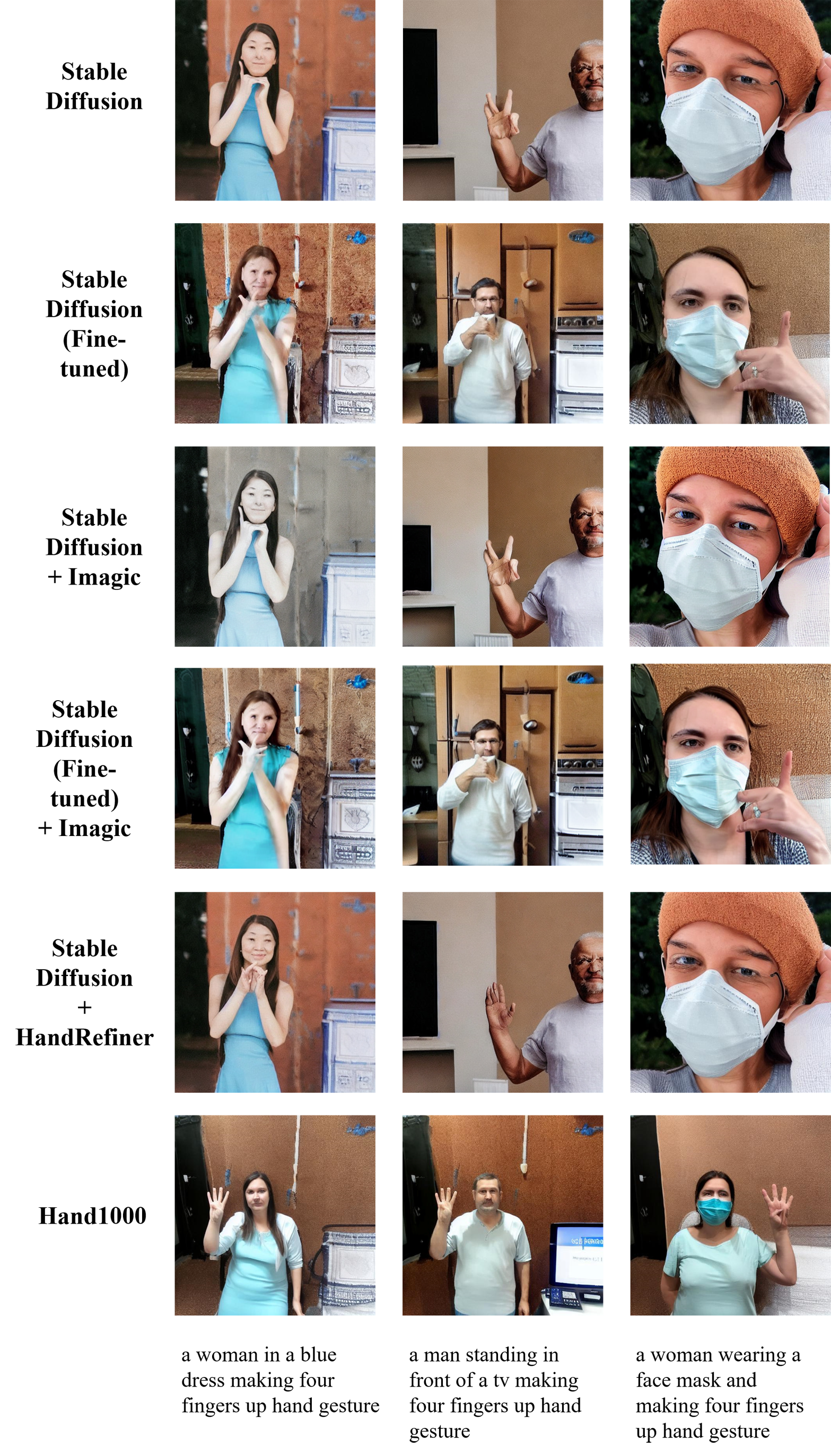}
   \caption{Comparison of images in hand gesture of four fingers up generated by stable diffusion, fine-tuned stable diffusion, stable diffusion enhanced with Imagic~\cite{kawar2023imagic}, fine-tuned stable diffusion enhanced with Imagic~\cite{kawar2023imagic}, stable diffusion enhanced with HandRefiner~\cite{lu2023handrefiner} and our Hand1000. }
   \label{fig:call_four}
\end{figure}


\subsection{Dataset Construction}
As no existing datasets specifically designed for text-to-hand image generation are publicly available, we construct a new dataset to bridge this gap. The process of the dataset curation is illustrated in Figure \ref{fig:dataset_fig}. 
Our dataset is built upon HaGRID~\cite{kapitanov2024hagrid}), which is a comprehensive collection of various hand gestures. This dataset features clear hand images and includes a diversity of individuals, clothing, and colors, making it highly suitable for our training and evaluation purposes. We construct our datasets using six hand gestures—``phone call," ``four," ``like," ``mute," ``ok," and ``palm" from HaGRID. 1,000 images are selected for each gesture for training and testing respectively. These hand gestures include ones with high finger separation (e.g., ``palm"), tightly closed fingers (e.g., ``mute"), and complex poses (e.g., ``ok"), providing a comprehensive assessment of the model's capabilities.

Nevertheless, HaGRID~\cite{kapitanov2024hagrid}) only contains hand gesture labels and lacks detailed textual descriptions of the images, which can not be directly used to train our model. 
To address this issue, we first adopt image captioning models to generate a textual description for each image. To inject hand gesture information into the textual description, we input the gesture labels and textual descriptions into the LLaMA3 model~\cite{touvron2023open}, enriching the original text with rich hand gesture information. We employ three different image captioning models—BLIP2~\cite{li2023blip}, PaliGemma~\cite{beyer2024paligemma}, and VitGpt2~\cite{mishra2024image}—to generate the textual descriptions. To evaluate the quality of the textual descriptions, we use CLIP's~\cite{radford2021learning} text and image encoders 
to calculate the similarity between the textual descriptions and corresponding images as well as the pairwise similarity between texts. As shown in Table \ref{tab:dataset}, the results show that the text-image similarity improved after postprocessing with LLaMA3~\cite{touvron2023open}, while the textual similarity does not increase significantly, demonstrating the effectiveness of this approach. According to Table \ref{tab:dataset}, BLIP2~\cite{li2023blip} shows the best performance, which is thus chosen as the image captioning model for generating textual descriptions for our dataset. 

\begin{table*}[t]
  \centering
  \caption{Dataset construction quality comparison using different image captioning models and LLaMA3 for postprocessing.}
    \small
    \begin{tabular}{lcc}
    \toprule
    Dataset Processing Method & Image-Caption Consistency↑ & Caption-Caption Similarity↓ \\
    \midrule
    PaliGemma & 0.237 & 0.835 \\
    VitGpt2 & \underline{0.254} & \underline{0.643} \\
    \textbf{BLIP2} & \textbf{0.288} & \textbf{0.600} \\
    \midrule
    PaliGemma(post-processed) & 0.252 & 0.715 \\
    VitGpt2(post-processed) & \underline{0.273} & \textbf{0.617} \\
    \textbf{BLIP2(post-processed)} & \textbf{0.298} & \underline{0.636} \\
    \bottomrule
    \end{tabular}%
  \label{tab:dataset}%
\end{table*}

\begin{table*}[t]
  \centering
  \caption{Performance comparison with Stable Diffusion, fine-tuned Stable Diffusion and HandRefiner. ↓ indicates that a lower value is better for that metric, while ↑ signifies that a higher value is better.}
    \small
    \begin{tabular}{lccccc}
    \toprule
    Method & FID↓  & KID↓  & FID-H↓ & KID-H↓ & HAND-CONF↑ \\
    \midrule
    Stable Diffusion & 114.768 & 0.079 & 109.168 & 0.094 & 0.895 \\
    Stable Diffusion(Fine-tuned) & \textbf{76.120} & \textbf{0.040} & 123.254 & 0.080 & 0.913 \\
    HandRefiner & 97.565 & 0.068 & \underline{88.997} & \underline{0.056} & \underline{0.919} \\
    \textbf{Hand1000} & \underline{86.169} & \underline{0.062} & \textbf{78.960} & \textbf{0.053} & \textbf{0.948} \\
    \bottomrule
    \end{tabular}%
  \label{tab:base}%
\end{table*}%

\subsection{Evaluation Metrics}

we employ five evaluate metrics to assess the quality of the generated images: Frechet Inception Distance (FID), Kernel Inception Distance (KID), FID-H, KID-H, and HAND-CONF~\cite{heusel2017gans, parmar2022aliased}. FID-H involves using an open-source hand gesture recognition model (i.e. Mediapipe hands~\cite{zhang2020mediapipe}) to segment the hand regions in the images, and then computing the FID on these segmented patches; KID-H is calculated similarly. The first four metrics measure the distance between the feature distributions of generated images and real images, with smaller distances indicating that the generated images are closer to real ones and thus of higher quality. HAND-CONF is computed by running a pre-trained hand detector(i.e. Mediapipe hands~\cite{zhang2020mediapipe}) and calculating the confidence scores of the detections. Higher confidence scores suggest that the detector is more certain about a region being a hand, indicating better quality of hand generation.

\subsection{Implementation Details}

For each image in the training set, the first stage of training involves 10 epochs with a learning rate of 1e-3. The second stage consists of 20 epochs with a learning rate of 1e-6. Besides, We use Mediapipe hands~\cite{zhang2020mediapipe} as the hand gesture recognition model while the text encoder is the CLIP~\cite{radford2021learning} text encoder. The sampling scheme is DDIM~\cite{song2020denoising} and the optimizer is Adam~\cite{kingma2014adam}. 
We used SDv1.4 for our experiments, which is similar to recent works like HanDiffuser~\cite{narasimhaswamy2024handiffuser}, HandRefiner~\cite{lu2023handrefiner}, and Imagic~\cite{kawar2023imagic}.

\subsection{Quantitative Performance Comparison}
As shown in Table \ref{tab:base}, our Hand1000 model demonstrates a significant performance improvement over the Stable Diffusion~\cite{rombach2022high} in terms of all the metrics. 
Specifically, Hand1000 achieves a 28.599 reduction in FID, a 0.017 reduction in KID, a 30.208 reduction in FID-H, and a 0.041 reduction in KID-H, while its HAND-CONF score is elevated by 0.053. To further validate the effectiveness of our method, we also perform comparison with fine-tuned stable diffusion using the same dataset. Although the fine-tuned Stable Diffusion model demonstrates slightly superior performance in terms of FID and KID compared to our method, several studies~\cite{kynkaanniemi2019improved, borji2019pros} have suggested that these two metrics are not entirely reliable. Nevertheless, the quality of hand image generation is significantly inferior, as indicated by FID-H, KID-H, and HAND-CONF. Compared to the fine-tuned stable diffusion model, our method achieves a 44.294 reduction in FID-H, a 0.027 reduction in KID-H, and a 0.035 increase in HAND-CONF, underscoring the effectiveness of our approach. Furthermore, we also conduct experiments on the open-source hand image refinement method, HandRefiner~\cite{lu2023handrefiner}, and the results demonstrate that our method outperforms HandRefiner~\cite{lu2023handrefiner} across all metrics, with an 11.396 reduction in FID, a 0.006 reduction in KID, a 10.037 reduction in FID-H, a 0.003 reduction in KID-H, and a 0.029 increase in HAND-CONF.

\subsection{Qualitative Example Comparison}
 By providing identical text prompts, we conduct a comparative analysis of the results with Stable Diffusion, finetuned Stable Diffusion, using Imagic~\cite{kawar2023imagic} to edit the results of Stable Diffusion and fine-tuned Stable Diffusion as well as HandRefiner. As illustrated in Figure \ref{fig:call_four}, our method not only generates realistic hand images but also effectively renders other components such as characters and clothing, demonstrating superior performance compared to all the other methods. 
 Furthermore, though Imagic is adopted to edit the image generated by Stable Diffusion, the result led to subtle alterations, such as the lightening of the dress color in the first set of images in the bottom row. In other words, Imagic fails to recover realistic hands and bodies from the generated images, leaving them in a chaotic, deformed, and distorted state. Additional experimental results can be found in the supplementary materials. Besides, HandRefiner fails to interpret gesture information accurately, despite some improvement in hand deformation, and shows no effect when only partial hands are present, as in the rightmost example.

\begin{figure}[t!]
  \centering
   \includegraphics[width=0.85\linewidth]{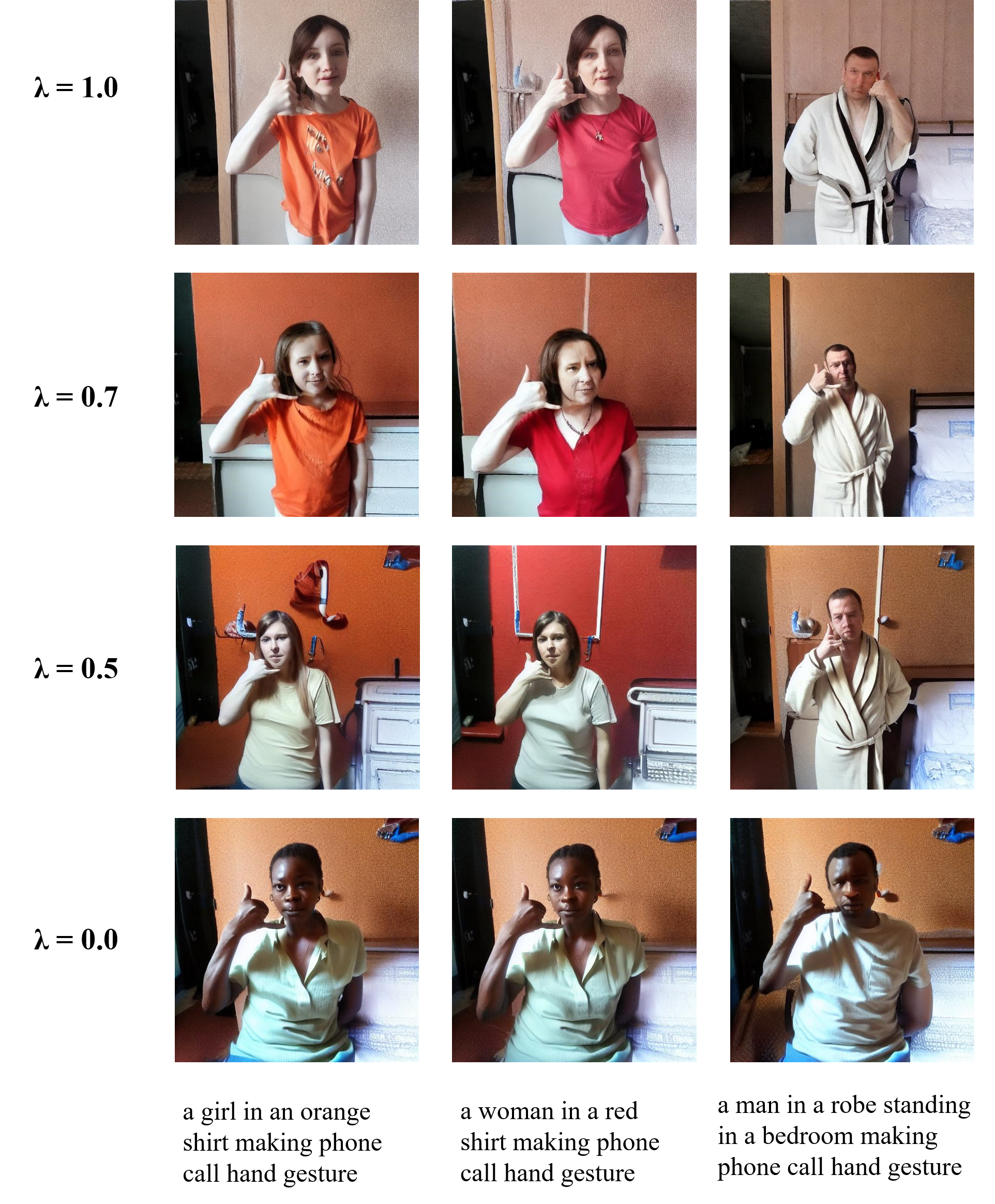}
   \caption{The effect of different values of $\lambda$ is evaluated. For the ``phone call'' gesture, the optimal value is found to be $\lambda=0.7$. When $\lambda$ is too large, the hand regions appear abnormal, whereas when $\lambda$ is too small, the generated images tend to become homogenized.}
   \label{fig:call_lambda}
\end{figure}

\subsection{Ablation Studies}
To validate the effectiveness of our design choices, we present a series of ablation experiments, including the ablation of various main components of the method and the ablation of several hyperparameters. 
\subsubsection{Effect of Embedding Optimization and Feature Fusion in Stage \setcounter{mycounter}{2}\Roman{mycounter}}
The ablation results for the feature fusion and embedding optimization in Stage \setcounter{mycounter}{2}\Roman{mycounter} are shown in Table \ref{tab:ablation}. Since FID-H and KID-H metrics are calculated after extracting the hand region, they provide a more accurate measure of the model's hand generation capability. With the addition of embedding optimization, FID-H is decreased by 25.003 and KID-H by 0.035. Incorporating Feature Fusion, which integrates features extracted by the hand gesture recognition model with text embeddings, leads to a reduction of 20.319 in FID-H and 0.048 in KID-H. When both techniques are used together, FID-H is decreased by 29.636 and KID-H by 0.051, demonstrating the superiority of our designs.

\subsubsection{The Impact of Hyperparameter $\lambda$ }

$\lambda$, as defined in Equation \ref{eq:eq2}, controls the proportion of the Text Embedding in the linear fusion step. A larger $\lambda$ indicates a higher proportion of the Text Embedding. The optimal value of $\lambda$ varies across different hand gestures and datasets. Accordingly, the parameter $\mu$ in Equation \ref{eq:eq3} must be adjusted during the inference phase to reflect changes in $\lambda$. Figure \ref{fig:call_lambda} illustrates the effects of varying $\lambda$ on the gesture ``phone call'' when $\mu$ is optimally tuned. It can be observed that when $\lambda$ is too low, the generated hand regions are accurate and well-formed, but other features, such as the person and clothing, tend to become homogenized. This is because a low $\lambda$ reduces the diversity of the Double Fused Embedding. Conversely, when $\lambda$ is too high, the diversity of the images is maintained, but the generated hand regions may exhibit distortions, as the hand feature fusion is not fully utilized. For the ``phone call'' gesture, an optimal value of $\lambda=0.7$ was found, which produces normal hand regions while maintaining diversity and realism in other parts of the image.

\subsubsection{Size of Training Set}

\begin{table}[t!]
  \centering
  \caption{Ablation Study of Embedding Optimization and Feature Fusion in Stage \setcounter{mycounter}{2}\Roman{mycounter}.}
    \small
    \begin{tabular}{cccc}
    \toprule
    Emb Optimization & Feature Fusion & FID-H↓ & KID-H↓ \\
    \midrule
    \textbf{\texttimes}     & \textbf{\texttimes}     & 108.596 & 0.104 \\
    \textbf{\texttimes}     & \textbf{\checkmark}     & 88.277 & 0.056 \\
    \textbf{\checkmark}     & \textbf{\texttimes}     & 83.593 & 0.069 \\
    \textbf{\checkmark}     & \textbf{\checkmark}     & \textbf{78.960} & \textbf{0.053} \\
    \bottomrule
    \end{tabular}%
  \label{tab:ablation}%
\end{table}%

\begin{figure}[t!]
  \centering
   \includegraphics[width=0.85\linewidth]{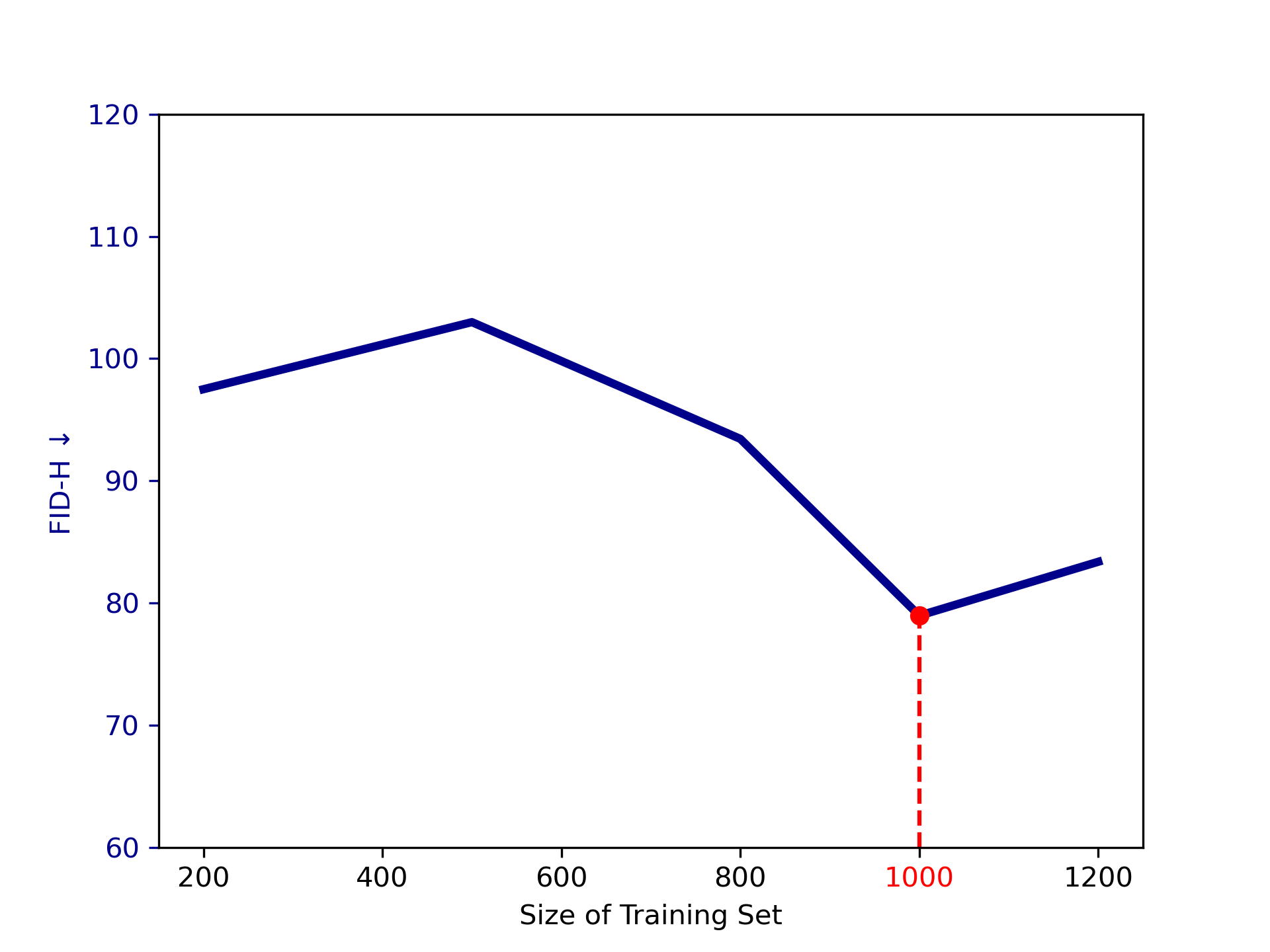}
   \caption{Ablation for the size of the training set. }
   \label{fig:FID-H}
\end{figure}

To determine the optimal training set size for our method, we conduct ablation experiments using FID-H as the evaluation metric. As shown in Figure \ref{fig:FID-H}, the experiments reveal that when the training set size is below 1,000, the FID-H is relatively high. The FID-H begins to decrease at a training set size of 500, reaching its minimum value at \textbf{1,000}. At a training set size of 1,200, the FID-H shows a slight increase, but the magnitude of this increase is small, which may indicate overfitting of the embedding, a known challenge in image-editing methods and worth further study. Therefore, we select {1,000} as the optimal training set size, and ultimately decided to use only {1,000} images for all experiments to ensure both efficiency and performance. 

\section{Conclusion}
\label{sec:Conclusion}

We have presented a novel method based on the diffusion model to generate realistic hand images from text using only 1,000 images. Through quantitative and qualitative experiments, we demonstrate the superiority and effectiveness of our approach. In particular, we show that it is significant to incorporate hand gesture information and optimize text embeddings for accurate hand image generation. As text-to-image generation continues to evolve, the methods and insights presented in this paper pave the way for more accurate and reliable visual representations, particularly in the challenging domain of human hands. While encouraging, this work is limited by the lack of exploration into simultaneous target gesture generation and hand-object interaction, primarily due to the unavailability of datasets, which will be our future work.

We would also like to gratefully acknowledge Bitdeer.AI for providing the cloud services and computing resources that were essential to this work.

\bibliography{main}

\end{document}